\documentclass[sn-mathphys]{sn-jnl}

\jyear{2021}%

\theoremstyle{thmstyleone}%

\theoremstyle{thmstyletwo}%

\theoremstyle{thmstylethree}%

\begin{document}

\title[ATTENTION GATE IN TRAFFIC FORECASTING]{ATTENTION GATE IN TRAFFIC FORECASTING}


\author*[1,2]{\spfx{Duc} \fnm{Anh}  \sur{Lam}}

\author[3]{\fnm{Anh}  \sur{Nguyen}}

\author[1,2]{ \sur{Bac} \fnm{Le}}

\affil*[1]{Faculty of Information Technology, University of Science, Ho Chi Minh City, Viet Nam}

\affil[2]{Vietnam National University, Ho Chi Minh City, Vietnam}

\affil[3]{Imperial College London, UK}


\abstract{
    Because of increased urban complexity and growing populations, more and more challenges about predicting city-wide mobility behavior are being organized.
    Traffic Map Movie Forecasting Challenge 2020 is secondly held in the competition track of the Thirty-fourth Conference on Neural Information Processing Systems (NeurIPS). 
    Similar to Traffic4Cast 2019, the task is to predict traffic flow volume, average speed in major directions on the geographical area of three big cities: Berlin, Istanbul, and Moscow. 
    In this paper, we apply the attention mechanism on U-Net based model, especially we add an attention gate on the skip-connection between contraction path and expansion path. An attention gates filter features from the contraction path before combining with features on the expansion path, it enables our model to reduce the effect of non-traffic region features and focus more on crucial region features.
    In addition to the competition data, we also propose two extra features which often affect traffic flow, that are time and weekdays.
    We experiment with our model on the competition dataset and reproduce the winner solution in the same environment. Overall, our model archives better performance than recent methods. 
}




\maketitle

\section{Introduction}
\label{sec1}

Nowadays, map is becoming more and more popular and is necessary for our life. It helps people figure out where they are, know the way to their destinations, and know how far they arrive there.
One of the important functions of the map is the function to calculate the travel time to the destination.
The calculation of the estimated time of arrival needs to be exact because people can rely on it to plan future works. The estimated time is measured not only depends on the distance between locations but also the traffic flow at a different times in a day. Traffic flow forecasting is an important problem in some practical applications.    

Traffic flow forecasting is also the main problem of Traffic Map Movie Forecasting Challenge 2020 (Traffic4cast 2020). The goal of the problem is to predict future traffic frames based on historical given traffic frames.
The provided time series probe data are the result of the combination of temporal and spatial aspects in this problem. 
Treating the problem as Frame Prediction task makes the problem become the fusion of image-to-image translation \cite{Liu2017-tm} and sequence-to-sequence \cite{Venugopalan2015-tt}.
At first glance, traffic flow casting is a time series problem and can be solved similarly in the work \cite{Nguyen2019-iy} by using popular backbone (ResNet \cite{He2016-zf}, VGG \cite{Simonyan2014-us}, Inception \cite{Szegedy2016-nl}) to extract features in each input frame, then feed them into Recurrent Neural Network (RNN) to help capture the temporal relationship between features.

Overall, almost all solutions for Traffic4Cast 2020 build their models based on U-Net architecture \cite{Ronneberger2015-cj}, which is the architecture of winner solutions in both versions Traffic4Cast challenge. These models preprocess input and output frames by concatenating them on channel axis and treat the problem like Image Segmentation problem. The difference between these models is the structure of the blocks on the contraction path and expansion path in U-Net architecture.
Moreover, with the above treatment, the work \cite{Wu2020-iw} try different Image Segmentation models to solve the problem.
To the best our knowledge, there is no U-shaped solution, which modifies the skip-connection between contraction path and expansion path.
Inspired by the first place solution to this year's competition, our model is also based on UNet architecture.

The contributions of our paper are summarized as follow:
(1) We propose a simple 2D attention gate based on contribution in the work \cite{Oktay2018-fj}. Our attention gate can be plugged in skip-connection in order to filter encoder features before it is passed into decoder block. 
(2) We propose two extra features, which often affect traffic flow prediction.
(3) We conduct two experiments: the first is one model for one city and the second is one model for three cities. We consider winner solution \cite{choi2020utilizing} to be a state-of-the-art model, and treat it as a baseline model. We experiment on the baseline model and our model in the same environment, and our model archive better performance on the first experiment and the second experiment.

\section{Data and Problem definition}
\label{sec2}

\subsection{Data}
\label{subsec:data}
The organizer provides 3 real-world data collected over the year 2019, which correspond to 3 entire cities: Berlin, Istanbul, and Moscow. Each data consists of training set, validation set, and test set.
The training set contains 181 files as 181 days of the first half-year from \date{January $1^{st}$ 2019} to \date{June $30^{th}$  2019}, the rest of the year data is in the test set. The validation set consists of 18 days sampled from the test set. In this competition, each city is divided into a $495 \times 436$ grid, a pixel of this grid represents a hundred square meters area of the city.
The data of each city has two types of data: dynamic data (e.g., traffic volume, traffic speed, incident level), and static data.
In the dynamic data, each day has 288 frames, each frame has nine channels, which show the aggregated information over a 5-minute period in a day. The channels include the traffic volume, average speed in each ordinal direction (e.g. North West, North East, South West, and South East), and encoding of recorded road incidents. Thereby each day is represented as a (288, 495, 436, 9) tensor. The data of volume, speed, incident level are normalized in the range [0, 255] through a max-min scaler. The missing value is equal to 0.
The difference from last year's competition is the static data for each city, which is a (495, 436, 7) tensor. The first two channels are a representation of the junction count while the next five channels encode the number of eating, drinking, and entertainment places, hospitals, parking places, shops, and public transport venues in that order. \newline

\subsection{Problem}
\label{subsec:problem}
The traffic flow forecasting problem in the Traffic4cast challenge is similar to Future Frame Prediction or Video Prediction, particularly to use the given 12 historical frames representing one-hour-long traffic map of cities and predict the 6 future frames for the next 5, 10, 15, 30, 45, 60 minutes respectively.

We formally define the problem as follows. Let $X_t \in \mathbf{R}^{495 \times 436 \times 9}$ be the $t$-th frame in the input frames sequence $\mathbf{X} = \{ X_{t - 11}, X_{t - 10}, \dots, X_t \}$ with twelve frames
where $495, 436$ are the height and width of frame, $9$ is a number of channels. The target is to predict the next frames $\mathbf{Y} = \{ \hat{Y}_{t + 1}, \hat{Y}_{t + 2}, \hat{Y}_{t + 3}, \hat{Y}_{t + 6}, \hat{Y}_{t + 9}, \hat{Y}_{t + 12}\}$.


\section{Related Work}
\label{sec3}

\textbf{Traffic Map Movie Forecasting}.
Through the solutions from the competition, we divide the approach into two main categories based on how the input data is processed: the first is to concatenate the input frames on channel axis, the second is to keep the input frames unchanged. With the first approach, the problem is treated as Image Segmentation problem, Models following first approach will receive an image as input, predict an image, and simultaneously the ground truth frames sequence is also 
concatenated on channel axis as input frames sequence and compared with the predicted image.
The most used architecture to solve the problem is U-Net architecture. In the work \cite{choi2019traffic,choi2020utilizing}, Sungbin Choi tried various U-Net based models by changing the structure of blocks on contraction path and expansion path. He consecutively won the first prize at the challenge in both times. Inspired by the winner solution of Traffic4cast 2019, most of the solutions \cite{Xu2020-zs,Bojesomo2020-ts} at the second Traffic4cast contest are variations of U-Net Model. Besides, some novel solutions approach this problem by using Graph Convolution in Graph Neural Network to capture Spatio-temporal features. Qi Qi et al. proposed a solution that combines Graph Neural Network with U-Net architecture.
With the second approach, the sequences of input frames remains the same, the type of model following this approach is usually the Recurrent Neural Network (RNN). Each input frame can be fed into a Convolutional Neural Network for feature extraction, then the extracted features are passed into a Recurrent Neural Network for sequential processing in order to predict the output frames. This method is used by \cite{Nguyen2019-iy}. The convolution blocks in the U-Net model can also be replaced by small recurrent neural networks to combine the power of the U-Net architecture with the sequential processing capabilities of the RNN, this method is proposed in the work \cite{Santokhi2020-ul}. In addition, to increase the model's performance, the work \cite{Wu2020-iw} has added features that can affect traffic such as day of the week, time of day, or Jingwei's method \cite{Xu2020-zs} constructs a mask that filters out the traffic-free areas to recap the predicted frames.

\noindent \textbf{Attention Mechanism}. Attention mechanisms originated from the investigations of human vision, only a part of all visible information is noticed by humans. Inspired by these attention visual mechanisms, researchers have tried to find a way how to apply attention to Deep Neural networks. Attention mechanisms enable a neural network to focus more on abundant informative elements of the input than on irrelevant parts. They were first introduced in Natural Language Processing (NLP), where encoder-decoder attention modules were developed to facilitate Neural Machine Translation \cite{bahdanau2016neural,nguyen2019object,Luong2015-co,gehring2017convolutional}. Trainable attention is enforced by design and categorized as hard attention and soft attention. Hard attention \cite{Mnih2014-nz}, e.g. iterative region proposal and cropping, is often not trained by back-propagation and relies on Reinforcement Learning for parameter updates, which makes the training phase more difficult and slow. By contrast, soft attention is probabilistic. It is more easily plugged in Deep Neural Networks and can be trained by standard back-propagation. For instance, additive soft attention is used in seq2seq translation \cite{bahdanau2016neural,nguyen2019v2cnet,Shen2018-zy} and recently applied to image classification \cite{Jetley2018-yi} and person identifications \cite{nguyen2021graph}. In the work \cite{Hu2018-gc}, channel-wise attention is used to highlight important feature dimensions, which has significantly top performance in the ILSVRC 2017 image classification challenge. Self-attention techniques \cite{Jetley2018-yi,Wang2018-oq} have been proposed to remove the dependency on external gating information.

\section{Methods}
\label{sec4}

\begin{figure*}
    \begin{center}
        \includegraphics[width=\textwidth]{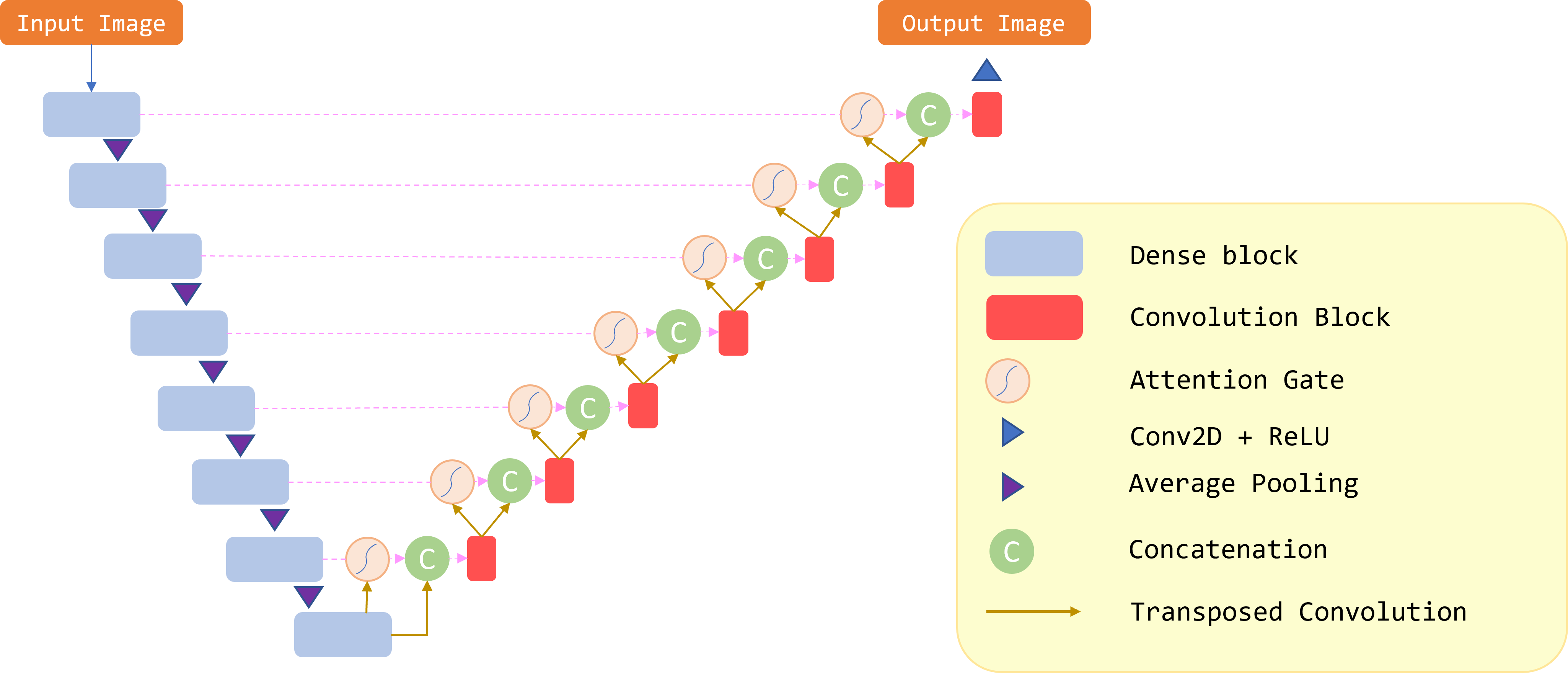}
    \end{center}
    \caption{The artchitecture of our U-Net based proposed model. In the training stage and validating stage, input frames are concatenated on channel axis in order to product only one input image, and the output frames is also concatenated. }
    \label{fig:overall_artchitecture}
\end{figure*}

\subsection{U-Net based model with Attention Gate}
Inspired by the work \cite{Oktay2018-fj}, we proposed a U-Net based model, which has the architecture as shown in figure \ref{fig:overall_artchitecture}. Similar to U-Net \cite{ronneberger2015unet}, our model consists of two symmetrical parts: contraction path and expansion path. 

The contraction path has consecutive dense blocks connected by the average pooling layer. Our dense block is the same as dense block in the work \cite{huang2018densely} but more simple. It has three stages connected together by dense connection. The first stage and second stage contain one convolution layer with $3 \times 3$ kernel size, ReLU activation layer, and one group normalization layer. We use group normalization instead of batch normalization because the batch size of input is too small due to batch normalization does not work well. The third stage contains only one convolution with $1 \times 1$ kernel size.
In the contraction path, the height and width of the image gradually reduce (downsampling, because of pooling) which helps the filters in the deeper layers to focus on a larger receptive field (context). However, the number of channels/depth (number of filters used) gradually increase which helps to extract more complex features from the image. By downsampling, the model better understands “WHAT” information is present in the image, but it loses the information of “WHERE” information it is present. 

After downsampling, the resolution of image is reduced. But
the output image must be the same resolution as the input image. Moreover, we use a regular convolutional network with pooling layers, we will lose the “WHERE” information and only retain the “WHAT” information which is not only what we want. In case of the problem, we need both “WHAT” as well as “WHERE” information. Hence there is a need to upsample the image, convert a low-resolution image to a high-resolution image to recover the “WHERE” information. That is also the target of the expansion path. The expansion path contains consecutive convolution blocks connected by the transposed convolution layer. The convolution block simply comprises of multiple convolution layers and pooling layers. Input of the convolution block is the combination of output of corresponding dense block on contraction path and output of preceding transposed convolution layer. The output of dense block is sent to the expansion path via skip-connection.

We propose an Attention Gate, which is added to each skip-connection. Before combining with features on the expansion path, the output of dense block on the contraction path
is fed into the attention gate in order to identify salient image regions and prune feature responses to preserve only the activations relevant to the specific task. We discuss more about attention gate in the section \ref{attention_gate_section}. 

\subsection{Attention Gate} \label{attention_gate_section}

In the traffic problem, a city will have traffic areas and no traffic areas. We found that using information in all areas would be redundant. Therefore, we propose Attention Gate with the aim to eliminate unimportant feature regions and allow important features containing useful information to pass through.

\begin{figure}
    \centering
    \includegraphics[width=\textwidth]{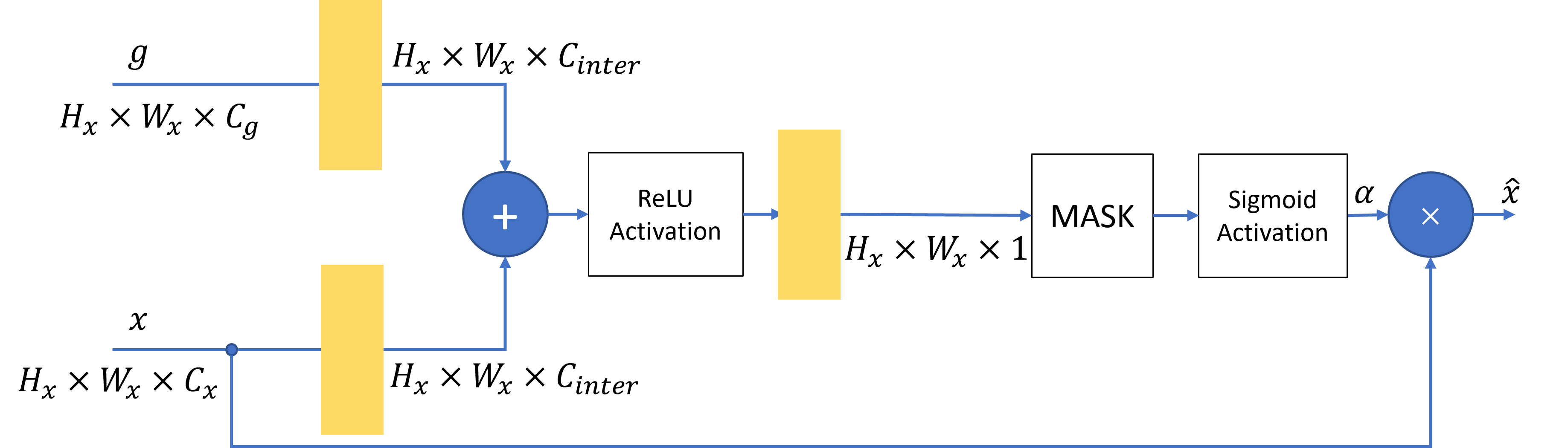}
    \caption{Structure of Attention Gate. The yellow block is convolution layer, which has $1 \times 1$ kernel size and no activation.}
    \label{fig:attention_gate}
\end{figure}

The structure of the Attention Gate is shown in figure \ref{fig:attention_gate}. The attention gate receives two inputs $x \in \mathbf{R}^{H_x \times W_x \times C_x}$ and $g \in \mathbf{R}^{H_x \times W_x \times C_g}$. $x$ is the output of the dense block on the contraction path. $g$ is the output of the transposed convolution layer at the lower stage. The gating vector contains contextual information to prune lower-level feature responses as suggested in \cite{Wang2017-kk}. Both $x$ and $g$ are fed into a convolution layer with $1 \times 1$ kernel size to make them have the same number of channels. Then they are combined by element-wise addition and fed into ReLU activation layer to perform non-linear operation. The output of ReLU passes through a convolution layer to make a mask having one channel. The mask is applied with a sigmoid function to normalize element values in the mask between 0 and 1. We call mask elements is attention coefficients. The coefficient value close to 1 means the feature is important, and vice versa. Finally, the output of the attention gate is the element-wise multiplication of input feature maps and attention coefficients.

\subsection{Extra Features}

\subsubsection{Weekdays feature}
\label{weekdays_feature}

Traffic flow is affected by days of week. On the workdays, traffic volume is usually higher than on the weekends, because of the traveling to work of commuters.
So, we propose a days-of-week feature. There are 7 days in a week, each day in week can be indexed from 0 to 6. For example, Monday is index 0, Tuesday is index 1, and so on. From the given day, we identify the weekday corresponding to that day, then we know the index of that day. We use one-hot encoding to convert each day of the dataset to a seven-element vector. The vector element, which has the same index as the index of the day we consider, is set to 1. Other elements of that vector are set to 0. With each input pixel, we concatenate the vector above into the data vector of the pixel on channel axis. To add all weekdays vector to every pixel of input frame, we build a tensor whose size is $495 \times 436 \times 7$.  For each channel of the recently built tensor, the elements of the channel, which has the index are the same as the index of the day considered in the week, is set to 1, and the others are set to 0. Then, we concatenate the tensor into the input frame.

\subsubsection{Time-a-day feature}
\label{timeaday_feature}

The time of day is one of the factors affecting traffic flow. For each input of the model, there are 12 times of day, which represent 12 inputs frames. The time of the first frame is chosen to be the time of that input. We have 288 frames so we index from 0 to 287. For example, the index of 00:00-00:05 is 0, the index of 23:55-00:00 is 1, so on and the index of 23:55-0:00 is indexed 287. Realizing that time is periodic so, we build a function $f_{time}(t)$ that maps time to coordinate of a point on the trigonometric unit circle. Function $f_{time}(t)$ is calculated as the Equation \ref{time_feature_formula}.

\begin{equation} \label{time_feature_formula}
        f_{time}(t) = ( \cos(\frac{t * 2 * \pi}{288}), \sin(\frac{t * 2 * \pi}{288}) ), 
        t \in \mathbf{N}, t \in [0, 287]
\end{equation}

We store that coordinate as a two-element vector, which we will concatenate the data vector of each pixel on channel axis. Similar to \ref{weekdays_feature}, we build a tensor whose size is  $495 \times 436 \times 2$. The value of elements at channel 0 equals  $\cos(\frac{t * 2 * \pi}{288})$ and the value of the elemets at channel 1 equals $\sin(\frac{t * 2 * \pi}{288})$. Then, we concatenate this tensor into the input frame in the dimension of data channel.

\section{Experiments}\label{sec5}

We train our proposed model on Google Colab \cite{ggcolab}.
Because limited training resource on Google Colab, we config an enviroment for each city as follow:
    \begin{itemize}
        \item Training set: 20 first data days taken from full challenge training set. Concretely, the days are from \date{January $1^{st}$ 2019} to \date{January $20^{th}$ 2019}.
        \item Validation set: 5 data days taken from full challenge validation set. Concretely, the days are \date{July $1^{st}$ 2019}, \date{July $11^{th}$ 2019}, \date{July $21^{st}$ 2019}, \date{July $31^{st}$ 2019}, \date{August $10^{th}$ 2019}. 
    \end{itemize}
    
We conduct three experiments. In the first experiment, we only use the dynamic data to train our proposed model and baseline model. In the second experiment, we use both the dynamic data and the static data to train the baseline model, our proposed model, and our proposed model using extra features. In the first and second experiments, we train one model for one city. The third experiment is the same as the second experiment but we train one model for all three cities.
We use Adam Optimizer for training models and set the learning rate as $3e-4$. The loss function of models is Mean Squared Error because the challenge relies on Mean Squared Error to rank the solution.
Results from experiments are reported below. 

\subsection{Experiment use only dynamic data - one model for one city}

In this experiment, we train a model from scratch for each city and evaluate it on the corresponding validation set. We only use dynamic data for training. The batch size is set to 2. The results of the experiment are showed in table \ref{table:exprberlin3}, table \ref{table:expristanbul3} and table \ref{table:exprmoscow3}. With a small training, our proposed model has a better performance in all three cities. This means the attention gates plugged on skip-connection work well and improve the performance of the model. 

\begin{table}[h]
\begin{center}
\begin{minipage}{210pt}
\caption{Result of first experiment on Berlin}
\label{table:exprberlin3}
\begin{tabular}{|l|c|c|}
          \hline
            \multicolumn{3}{|c|}{Berlin} \\
          \hline
              Model & Step & Mean Squared Error \\
          \hline\hline
              Baseline model &  1000 & 0.001496935 \\
              & 2000 & 0.001453448 \\
              & 2650 & 0.001424915 \\
          \hline
              Our proposed model &  1000 & 0.001485584 \\
              & 2000 & 0.001439420 \\
              & 2650 & \textbf{0.001415349} \\
          \hline
\end{tabular}

\end{minipage}
\end{center}
\end{table}

\begin{table}[h]
    \begin{center}
        \begin{minipage}{210pt}
        \caption{Result of first experiment on Istanbul}
        \label{table:expristanbul3}
        \begin{tabular}{|l|c|c|}
          \hline
            \multicolumn{3}{|c|}{Istanbul} \\
          \hline
              Model & Step & Mean Squared Error \\
          \hline\hline
              Baseline model &  1000 & 0.0011355047 \\
              & 2000 & 0.0010781622 \\
              & 2650 & 0.001057595\\
          \hline
              Our propose model &  1000 & 0.001104628 \\
              & 2000 & 0.001066083 \\
              & 2650 & \textbf{0.001050306} \\
          \hline
        \end{tabular}
        \end{minipage}
    \end{center}
\end{table}

\begin{table}[h]
    \begin{center}
        \begin{minipage}{210pt}
        \caption{Result of first experiment on Moscow}
        \label{table:exprmoscow3}
        \begin{tabular}{|l|c|c|}
          \hline
            \multicolumn{3}{|c|}{Moscow} \\
          \hline
              Model & Step & Mean Squared Error \\
          \hline\hline
              Baseline model &  1000 & 0.001612101 \\
              & 2000 & 0.001546476 \\
              & 2650 & 0.001516563 \\
          \hline
              Our proposed model &  1000 & 0.001581952 \\
              & 2000 & 0.001524919 \\
              & 2650 & \textbf{0.001416972} \\
          \hline
        \end{tabular} 
        \end{minipage}
    \end{center}
\end{table}

\subsection{Experiment use dynamic data and static data - one model for one city}

Similar to the first experiment, we train a model scratch for each city, but in the second experiment, we use both dynamic data and static data. We also supply proposed extra features to our proposed model. The results of the second experiment are showed in table \ref{table:exprberlin}, table \ref{table:expristanbul} and table \ref{table:exprmoscow}. From the result, our model archives better result in all three cities. The result of the second experiment is better than the one of the first experiment, which shows static data is necessary for more precise traffic flow forecasting. Moreover, our model using extra features outperforms the rest of the models. This means our extra features contribute to improving the accuracy of the model. 

\begin{table}[h]
    \begin{center}
        \begin{minipage}{210pt}
        \caption{Result of second experiment on Berlin}
        \label{table:exprberlin}
        \begin{tabular}{|l|c|c|}
          \hline
            \multicolumn{3}{|c|}{Berlin} \\
          \hline
              Model & Step & Mean Squared Error \\
          \hline\hline
              Baseline &  1000 & 0.001496405 \\
              & 2000 & 0.001437678 \\
              & 2650 & 0.001421472 \\
          \hline
              Our proposed model &  1000 & 0.0014768082 \\
              & 2000 & 0.0014298414 \\
              & 2650 & 0.0014169723 \\
          \hline
              Our proposed model  & 1000 & 0.0014784151 \\
              with extra features & 2000 & 0.0014241398 \\
              & 2650 & \textbf{0.00140963} \\
          \hline
        \end{tabular} 
        \end{minipage}
    \end{center}
\end{table}

\begin{table}[h]
    \begin{center}
        \begin{minipage}{210pt}
        \caption{Result of second experiment on Istanbul}
        \label{table:expristanbul}
        \begin{tabular}{|l|c|c|}
          \hline
            \multicolumn{3}{|c|}{Istanbul} \\
          \hline
              Model & Step & Mean Squared Error \\
          \hline\hline
             Baseline model &  1000 & 0.0011224896 \\
              & 2000 & 0.0010650043 \\
              & 2650 & 0.0010501256\\
          \hline
              Our proposed model &  1000 & 0.0010910966 \\
              & 2000 & 0.0010433174 \\
              & 2650 & 0.0010324720\\
          \hline
              Our proposed model  &  1000 & 0.0011083923 \\
              with extra features & 2000 & 0.0010441822 \\
              & 2650 & \textbf{0.0010284710} \\
          \hline
        \end{tabular}
        \end{minipage}
    \end{center}
\end{table}

\begin{table}[h]
    \begin{center}
        \begin{minipage}{210pt}
        \caption{Result of second experiment on Moscow}
        \label{table:exprmoscow}
        \begin{tabular}{|l|c|c|}
          \hline
            \multicolumn{3}{|c|}{Moscow} \\
          \hline
              Model & Step & Mean Squared Error \\
          \hline\hline
              Baseline model &  1000 & 0.0015991237 \\
              & 2000 & 0.0015390549 \\
              & 2650 & 0.0015111139 \\
          \hline
              Our proposed model &  1000 & 0.0016474014 \\
              & 2000 & 0.0015251556 \\
              & 2650 & 0.0014169724 \\
          \hline
              Our proposed model &  1000 & 0.0015787245 \\
              with extra features & 2000 & 0.0015262594 \\
              & 2650 & \textbf{0.0014096299} \\
          \hline
        \end{tabular} 
        \end{minipage}
    \end{center}
\end{table}

\subsection{Experiment use dynamic data and static data - one model for three cities}

In the third experiment, we try training one model on the training set of all three cities. Then we evaluate trained model on validation set of each city. The results of the third experiment are showed in table \ref{table:expr2_result}. The result is not what we expected. The mean square error of baseline model is lower than our models. We can explain the difference between third experiment and both previous experiments. Each city has a separate traffic network, so the region need to be focus is different. Applying attention gate on three cities will force the attention gate to focus important regions on three cities. The important region from one city can be not important in other cities, this leads to reduce the accuracy of attention gate. 

\begin{table}[h]
    \begin{center}
        \begin{minipage}{250pt}
        \caption{Result of third experiment on three cities}
        \label{table:expr2_result}
        \begin{tabular}{|c|l|c|}
          \hline
              City & Model & Mean Squared Error \\
          \hline\hline
              Berlin &  Baseline model  & \textbf{0.0014484838} \\
              & Our proposed model & 0.0015388483\\
              & \vtop{\hbox{\strut Our proposed model}\hbox{\strut with extra features}} & 0.0016585335 \\
          \hline
              Istanbul & Baseline model  & \textbf{0.0010459442} \\
              & Our proposed model & 0.0011944101 \\
              & \vtop{\hbox{\strut Our proposed model}\hbox{\strut with extra features}} & 0.001326789 \\
          \hline
              Moscow & Baesline model  & 0.0014623589 \\
              & Our proposed model & 0.0014423244 \\
              & \vtop{\hbox{\strut Our proposed model}\hbox{\strut with extra features}} & \textbf{0.0014409719} \\
          \hline
        \end{tabular}
        \end{minipage}
    \end{center}
\end{table}

\section{Conclusion}
\label{sec6}

Traffic flow forecasting is an important problem in many Intelligent Transportation Systems. Representing traffic data in the form of images opens up a potential approach. Existed Convolutional Neural networks can be leveraged to solve the problem. In this paper, we apply attention mechanisms into our proposed model to select the regions that need to focus on and eliminate the effect of unimportant regions. In addition to the data provided by the organizers, we add two more traffic-related features. Additional features have contributed to increasing the accuracy of the model. From there, we can select other traffic-influenced features to add to increase the performance of the model.

\bibliography{sn-bibliography}

\end{document}